\documentclass[10pt,twocolumn,letterpaper]{article}

\usepackage{multirow}
\usepackage{cvpr}
\usepackage{times}
\usepackage{epsfig}
\usepackage{graphicx}
\usepackage{amsmath}
\usepackage{amssymb}
\usepackage[export]{adjustbox}
\usepackage[numbers,sort]{natbib} 

\usepackage{multido}
\usepackage{subcaption}
\usepackage{comment}

\usepackage[pagebackref=true,breaklinks=true,letterpaper=true,colorlinks,bookmarks=false]{hyperref}

\cvprfinalcopy

\ifcvprfinal\fi
\begin{document}

\title{A Short Note about Kinetics-600}

\author{{Jo\~a}o Carreira \\
{\tt\small joaoluis@google.com} \\
\and
Eric Noland \\
{\tt\small enoland@google.com} \\
\and
Andras Banki-Horvath \\ 
{\tt\small bhandras@google.com} \\
\and
Chloe Hillier \\
{\tt\small chillier@google.com} \\
\and
Andrew Zisserman \\
{\tt\small zisserman@google.com} \\
}

\maketitle

\begin{abstract}

We describe an extension of the DeepMind Kinetics human action dataset from 400 classes, each with at least 400 video clips, to 600 classes, each with at least 600 video clips. In order to scale up the dataset we changed the data collection process so it uses  multiple queries per class, with some of them in a language other than english -- portuguese. This paper details the changes between the two versions of the dataset and includes a comprehensive set of statistics of the new version as well as baseline  results using the I3D neural network architecture. The paper is a companion to the release of the ground truth labels for the public test set.
\end{abstract}

\section{Introduction}

The release of the Kinetics dataset \cite{kay2017kinetics} in 2017 led to marked improvements in state-of-the-art performance on a variety of action recognition datasets: UCF-101 \cite{soomro2012ucf101}, HMDB-51 \cite{Kuehne11}, Charades \cite{sigurdsson2016hollywood}, AVA \cite{gu2017ava}, Thumos \cite{jiang2014thumos}, among others. Video models pre-trained on Kinetics generalized well when transferred to different video tasks on smaller video datasets, similar to what happened to image classifiers trained on ImageNet.

The goal of the Kinetics project from the start was to replicate the size of ImageNet, which has 1000 classes, each with 1000 image examples.  This proved difficult initially and the first version of the dataset had 400 classes, each with 400 video clip examples. There were two main bottlenecks and they were related: (a) identifying relevant candidate YouTube videos for each action class, and (b) finding classes having many candidates. Problem (b) was particularly acute and exposed inefficiencies with the way videos were selected -- querying YouTube for simple variations of the class names, by varying singular/plural of nouns, adding articles (e.g.\ ``catching a ball" / ``catching ball"), etc. These problems have now been overcome, as described in the sequel.

The new version of the dataset, called Kinetics-600, follows the same principles as Kinetics-400:
(i) The clips are from YouTube video, last 10s,  and have a variable resolution and frame rate; (ii) for an action class, all clips are from different YouTube videos.
 Kinetics-600 represents a 50\% increase in number of classes, from 400 to 600,  and a 60\% increase in the number of video clips,
from around 300k to around 500k. The statistics of the two dataset versions are detailed in table~\ref{tab:kinetics_stats}.

In the new Kinetics-600 dataset there is a standard test set, for which labels have been publicly released, and also
 a held-out test set (where the labels are not released).
We encourage researchers to report results on the standard test set, unless they want to compare with participants of the Activity-Net kinetics challenge. Performance on the combination of standard test set plus held-out test should be used in that case, and can be be measured only through the challenge evaluation website\footnote{http://activity-net.org/challenges/2018/evaluation.html}.

The URLs of the YouTube videos and temporal intervals of both Kinetics-600 and Kinetics-400 can be obtained from \url{http://deepmind.com/kinetics}.

\begin{table*}[ht]
\centering
\begin{tabular}{| l || c | c | c | c || c | c | c | }
  \hline
  \textbf{Version} & \textbf{Train} & \textbf{Valid.} & \textbf{Test} & \textbf{Held-out Test} & \textbf{Total Train} & \textbf{Total} & {\textbf{Classes}}  \\ \hline 
Kinetics-400 \cite{kay2017kinetics} & 250--1000 & 50  & 100 & 0  & 246,245 & 306,245 & 400 \\ \hline
Kinetics-600 & 450--1000 & 50  & 100 & around 50 & 392,622 & 495,547 & 600 \\  \hline
\end{tabular} 
\vspace{5pt}
\caption{Kinetics Dataset Statistics. The number of clips for each class in the various splits (left), and the totals (right). With Kinetics-600 we have released the ground truth test set labels, and also created an additional held-out test set for the purpose of the Activity-Net Challenge. }
\label{tab:kinetics_stats}
\end{table*}

\section{Data Collection Process \label{collection}}

The data collection process evolved from Kinetics-400 to Kinetics-600. The overall pipeline was the same: 1) action class sourcing, 2) candidate video matching, 3) candidate clip selection, 4) human verification, 5) quality analysis and filtering. In words, a list of class names is created, then a list of candidate YouTube URLs is obtained for each class name, and candidate 10s clips are sampled from the videos. These clips are sent to humans in Mechanical Turk who decide whether those clips contain the action class that they are supposed to. Finally, there is an overall curation process including clip de-duplication, and selecting the higher quality classes and clips. Full details can be found in the original publication~\cite{kay2017kinetics}. 

The main differences in the data collection process between Kinetics-400 and 600 were in the first two steps: how action classes were sourced, and how candidate YouTube videos were matched with classes.

\vspace{3mm}
\subsection{Action class sourcing} 
For Kinetics-400, class names were first sourced from existing
datasets, then from the everyday experience of the authors, and
finally by asking the humans in Mechanical Turk what classes they
were seeing in videos that did not contain the classes being tested.
For Kinetics-600 we sourced many classes from
Google's Knowledge Graph, in particular from the hobby list. We also
obtained class ideas from YouTube's search box auto-complete, for
example by typing an object or verb, then following up on promising
auto-completion suggestions and checking if there were many videos
containing the same action.

\vspace{3mm}
\subsection{Candidate video matching} 
In Kinetics-400 we matched YouTube videos with each class by searching for videos having some of the class name words in the title, while allowing for variation in stemming. There was no separation between the class name and the query text, which turned out to be a limiting factor: in many cases we exhausted the pool of candidates, or had impractically low yields. We tried matching directly these queries to not just the title but also other metadata but this proved of little use (in particular the video descriptions seemed to have plenty of spam). We tried two variations that worked out much better:

\vspace{3mm}
\noindent \textbf{Multiple queries.} 
In order to get better and larger pools of candidates we found it
useful to manually create sets of queries for each class and did so in
two different languages: English and Portuguese. These are two out of
six languages with the most native speakers in the
world\footnote{According to
https://www.babbel.com/en/magazine/the-10-most-spoken-languages-in-the-world/},
have large YouTube communities (especially in the USA and Brazil), and
were also natively spoken by this paper's authors.
As an example the queries for folding paper were: ``folding paper"
(en), ``origami" (en) and ``dobrar papel" (pt). We found also that
translating action descriptions was not always easy, and sometimes
required observing the videos returned by putative translated queries
on YouTube and tuning them through some trial and error.

Having multiple languages had the positive side effect of also promoting greater dataset diversity by incorporating a more well-rounded range of cultures, ethnicities and geographies.

\vspace{3mm}
\noindent \textbf{Weighted ngram matching.} Rather than matching directly using textual queries we found it beneficial to use weighted ngram representations of the combination of the metadata of each video and the titles of related ones. Importantly, these representations were compatible with multiple languages. We combined this with standard title matching to get a robust similarity score between a query and all YouTube videos, which, unlike the binary matching we used before, meant we never ran out of candidates, although the post-mechanical-turk yield of the selected candidates became lower for smaller similarity values.

\section{From Kinetics-400 to Kinetics-600}

Kinetics-600 is an approximate superset of Kinetics-400 -- overall,
368 of the original 400 classes are exactly the same in
Kinetics-600 (except they have more examples). For the other 32
classes, we renamed a few (e.g.\ ``dying hair" became ``dyeing hair"),
split or removed others that were too strongly overlapping with other
classes, such as ``drinking". We split some classes: ``hugging" became
``hugging baby" and ``hugging (not baby)", while ``opening bottle" became
``opening wine bottle" and ``opening bottle (not wine)".

A few video clips from 30 classes of the Kinetics-400 validation set
became part of the Kinetics-600 test set, and some from the
training set became part of the new validation set. 
It is
therefore not ideal to evaluate  models on Kinetics-600 that were
pre-trained on Kinetics-400, although it should make almost no
difference in practice. The full list of new classes in Kinetics-600 is given in
the appendix.

\section{Benchmark Performance}

\begin{table*}
\centering
\begin{tabular}{| c| r | r | c |}
  \hline
  \textbf{Acc. type} & \textbf{Valid} & \textbf{Test} & \textbf{Test + HeldOut Test}  \\ \hline
Top-1 & 71.9 & 71.7 & 69.7  \\ \hline
Top-5 & 90.1 & 90.4 & 89.1  \\ \hline
$100.0-avg$(Top-1,Top-5) & 19.0 & 19.0 & 20.6 \\ \hline
\end{tabular} 
\vspace{5pt}
\caption{Performance of an I3D model with RGB inputs on the Kinetics-600 dataset, without any test time augmentation (processing a center crop of each video convolutionally in time ). The first two rows show accuracy in percentage, the last one shows the metric used at the Kinetics challenge hosted by the ActivityNet workshop.}
\label{tab:benchmark}
\end{table*}

As a baseline model we used I3D \cite{Carreira17}, with standard RGB videos as input (no optical flow). 
We trained the model from scratch on the Kinetics-600 training set, picked hyper-parameters on validation, and report performance on validation, test set and the combination of the test and held-out test sets. We used 32 P100 GPUs, batch size 5 videos, 64 frame clips for training and 251 frames for testing. We trained using SGD with momentum, starting with a learning rate of 0.1, decreasing it by a factor of 10 when the loss saturates. Results are shown in table~\ref{tab:benchmark}.

The top-1 accuracy on the test set was 71.7, whereas on Test+Held-out was 69.7, which shows that the held-out test set is harder than the regular test set. On Kinetics-400 the corresponding result was 68.4, hence the task overall seems to have became slightly easier. There are several factors that may help explain this: 
even though Kinetics-600 has 50\% extra classes, it also has around 50\% extra training examples; and also, 
some of the ambiguities in Kinetics-400 have been removed in Kinetics-600. 
We also used fewer GPUs (32 instead 64), which resulted in half the batch size.

\vspace{3mm}
\noindent \textbf{Kinetics challenge.} 
There was a first Kinetics challenge at the ActivityNet workshop in
CVPR 2017, using Kinetics-400. The second challenge occurred at the
ActivityNet workshop in CVPR 2018, this time using Kinetics-600. The
performance criterion used in the challenge is the average of Top-1
and Top-5 error. There was an  improvement between the winning systems
of the two challenges, with error going down from 12.4\% (in 2017) to 11.0\% 
(in 2018)~\cite{bian2017revisiting,he2018exploiting}.

\section{Conclusion}
\label{sec:conclusion}

We have described the new Kinetics-600 dataset, which is 50\% larger
than the original Kinetics-400 dataset.  It represents another step
towards our goal of producing an action classification dataset with
1000 classes and 1000 video clips for each class. We explained the differences in
the data collection process between the initial version of the dataset
made available in 2017 and the new one. This publication coincides
with the release of the test set annotations for both Kinetics-400 and
Kinetics-600; we hope these will facilitate research as it will no
longer be necessary to submit results to an external evaluation
server.

\subsection*{Acknowledgements:} The collection of this dataset was funded by DeepMind. The authors would like to thank Sandra Portugues for helping to translate queries from English to Portuguese, and Aditya Zisserman and Radhika Desikan for data clean up.

{\small
\bibliographystyle{ieee}
\bibliography{references}
}

\appendix

\section{List of New Human Action Classes in Kinetics-600}
This is the list of classes in Kinetics-600 that were not in Kinetics-400,  or that have been renamed.

\begin{enumerate}
\itemsep0em 
\item acting in play
\item adjusting glasses
\item alligator wrestling
\item archaeological excavation
\item arguing
\item assembling bicycle
\item attending conference
\item backflip (human)
\item base jumping
\item bathing dog
\item battle rope training
\item blowdrying hair
\item blowing bubble gum
\item bodysurfing
\item bottling
\item bouncing on bouncy castle
\item breaking boards
\item breathing fire
\item building lego
\item building sandcastle
\item bull fighting
\item bulldozing
\item burping
\item calculating
\item calligraphy
\item capsizing
\item card stacking
\item card throwing
\item carving ice
\item casting fishing line
\item changing gear in car
\item changing wheel (not on bike)
\item chewing gum
\item chiseling stone
\item chiseling wood
\item chopping meat
\item chopping vegetables
\item clam digging
\item coloring in
\item combing hair
\item contorting
\item cooking sausages (not on barbeque)
\item cooking scallops
\item cosplaying
\item cracking back
\item cracking knuckles
\item crossing eyes
\item cumbia
\item curling (sport)
\item cutting apple
\item cutting orange
\item delivering mail
\item directing traffic
\item docking boat
\item doing jigsaw puzzle
\item drooling
\item dumpster diving
\item dyeing eyebrows
\item dyeing hair
\item embroidering
\item falling off bike
\item falling off chair
\item fencing (sport)
\item fidgeting
\item fixing bicycle
\item flint knapping
\item fly tying
\item geocaching
\item getting a piercing
\item gold panning
\item gospel singing in church
\item hand washing clothes
\item head stand
\item historical reenactment
\item home roasting coffee
\item huddling
\item hugging (not baby)
\item hugging baby
\item ice swimming
\item inflating balloons
\item installing carpet
\item ironing hair
\item jaywalking
\item jumping bicycle
\item jumping jacks
\item karaoke
\item land sailing
\item lawn mower racing
\item laying concrete
\item laying stone
\item laying tiles
\item leatherworking
\item licking
\item lifting hat
\item lighting fire
\item lock picking
\item longboarding
\item looking at phone
\item luge
\item making balloon shapes
\item making bubbles
\item making cheese
\item making horseshoes
\item making paper aeroplanes
\item making the bed
\item marriage proposal
\item massaging neck
\item moon walking
\item mosh pit dancing
\item mountain climber (exercise)
\item mushroom foraging
\item needle felting
\item opening bottle (not wine)
\item opening door
\item opening refrigerator
\item opening wine bottle
\item packing
\item passing american football (not in game)
\item passing soccer ball
\item person collecting garbage
\item photobombing
\item photocopying
\item pillow fight
\item pinching
\item pirouetting
\item planing wood
\item playing beer pong
\item playing blackjack
\item playing darts
\item playing dominoes
\item playing field hockey
\item playing gong
\item playing hand clapping games
\item playing laser tag
\item playing lute
\item playing maracas
\item playing marbles
\item playing netball
\item playing ocarina
\item playing pan pipes
\item playing pinball
\item playing ping pong
\item playing polo
\item playing rubiks cube
\item playing scrabble
\item playing with trains
\item poking bellybutton
\item polishing metal
\item popping balloons
\item pouring beer
\item preparing salad
\item pushing wheelbarrow
\item putting in contact lenses
\item putting on eyeliner
\item putting on foundation
\item putting on lipstick
\item putting on mascara
\item putting on sari
\item putting on shoes
\item raising eyebrows
\item repairing puncture
\item riding snow blower
\item roasting marshmallows
\item roasting pig
\item rolling pastry
\item rope pushdown
\item sausage making
\item sawing wood
\item scrapbooking
\item scrubbing face
\item separating eggs
\item sewing
\item shaping bread dough
\item shining flashlight
\item shopping
\item shucking oysters
\item shuffling feet
\item sipping cup
\item skiing mono
\item skipping stone
\item sleeping
\item smashing
\item smelling feet
\item smoking pipe
\item spelunking
\item square dancing
\item standing on hands
\item staring
\item steer roping
\item sucking lolly
\item swimming front crawl
\item swinging baseball bat
\item sword swallowing
\item tackling
\item tagging graffiti
\item talking on cell phone
\item tasting wine
\item threading needle
\item throwing ball (not baseball or American football)
\item throwing knife
\item throwing snowballs
\item throwing tantrum
\item throwing water balloon
\item tie dying
\item tightrope walking
\item tiptoeing
\item trimming shrubs
\item twiddling fingers
\item tying necktie
\item tying shoe laces
\item using a microscope
\item using a paint roller
\item using a power drill
\item using a sledge hammer
\item using a wrench
\item using atm
\item using bagging machine
\item using circular saw
\item using inhaler
\item using puppets
\item vacuuming floor
\item visiting the zoo
\item wading through mud
\item wading through water
\item waking up
\item walking through snow
\item watching tv
\item waving hand
\item weaving fabric
\item winking
\item wood burning (art)
\item yarn spinning

\end{enumerate}

\end{document}